\documentclass{article}
\usepackage{spconf,amsmath,amssymb,graphicx,booktabs,multirow,array,stfloats,float}
\usepackage[table]{xcolor}
\usepackage[hidelinks]{hyperref}
\usepackage[slantedGreek]{mathptmx}
\DeclareSymbolFont{operators}{OT1}{ptm}{m}{n}
\usepackage[T1]{fontenc}

\makeatletter
\renewcommand\section{\@startsection{section}{1}{\z@}%
  {-1.6ex \@plus -0.5ex \@minus -.2ex}%
  {1.0ex \@plus .2ex}{\normalfont\normalsize\bfseries}}
\renewcommand\subsection{\@startsection{subsection}{2}{\z@}%
  {-1.5ex \@plus 0pt \@minus 0pt}%
  {0.7ex \@plus 0pt}{\normalfont\normalsize\bfseries}}
\makeatother
\makeatletter
\long\def\@makecaption#1#2{%
 \vskip 4pt
 \setbox\@tempboxa\hbox{#1. #2}%
 \ifdim \wd\@tempboxa >\hsize #1. #2\par \else \hbox
to\hsize{\hfil\box\@tempboxa\hfil}%
 \fi}
\makeatother

\newcommand{\Dice}{\mathrm{Dice}}
\newcommand{\NA}{\multicolumn{1}{c}{--}}

\title{SetPlanner: A Lightweight Plug-in Point-Set Planner for Frozen SAM}

\name{Dawei Yan$^{1}$, Yuezhe Yang$^{2}$, Menglan Ruan$^{1}$, Chunfeng Yang$^{1}$, and Yudong Zhang$^{1}$\thanks{\small Yudong Zhang is the corresponding author.}}
\address{\small $^{1}$School of Computer Science and Engineering, Southeast Univ., China \quad $^{2}$School of Computer Science, Univ. of Sydney, Australia \\
\small \{220255428, 230250010, yudongzhang, chunfeng.yang\}@seu.edu.cn, yangyuezhe@gmail.com \\
\small Code: \url{https://github.com/davidyan200012-bot/SetPlanner}}

\let\sfoldbib\thebibliography
\renewcommand{\thebibliography}[1]{\sfoldbib{#1}\setlength{\itemsep}{0pt}\setlength{\parsep}{0pt}\setlength{\parskip}{0pt}\setlength{\baselineskip}{10.4pt}}

\begin{document}
\maketitle

  \begin{abstract}
  \small
  Segment Anything Models provide reusable priors, yet they require user prompts
  and cannot support fully automatic instrument segmentation.
  Automatic prompting is difficult for thin, articulated, reflective, and partly occluded
  tools, where several configurations can be valid. We formulate automatic
  prompting as lightweight point-set planning and isolate the point source under a frozen
  pathway. To this end, we present
  \textbf{SetPlanner}, a 1.52M-parameter
  plug-in point-set planner for frozen SAM. The plug-in preserves SAM's
  point-prompt interface and enables reuse across backbones. SetPlanner plans complete unordered $K$-point
  sets from geometry-aware targets with a permutation-aware
  conditional flow. SAM decodes eight candidates; their consensus readout yields a
  ground-truth-free prediction. Across three endoscopic datasets,
  SetPlanner wins all six transfer routes over a LoRA-adapted system. Under our
  frozen-pathway protocol, SetPlanner reaches 0.934 Dice on Kvasir-Instrument and
  recovers 96\% of a 44.4-point localization gap, while candidate disagreement ranks
  low-Dice cases at AUROC 0.969.
  \end{abstract}

\begin{keywords}
\small
Surgical instrument segmentation, SAM, automatic prompting, point-set
generation, flow matching
\end{keywords}

\begin{figure}[t]
\centering
\includegraphics[width=\columnwidth,trim=0 7.0bp 0 6.0bp,clip]{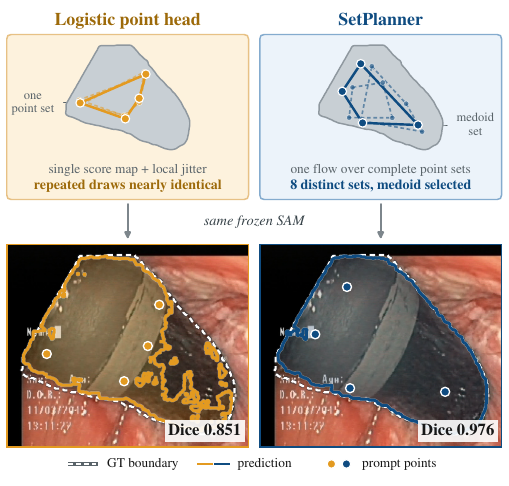}
\caption{\small \textbf{Point-source contrast.} Left: a single score map with local
jitter. Right: one flow over complete point sets. Each polygon links one
$K\!=\!4$ point set. The qualitative pair uses an official Kvasir-Instrument
test frame.}
\label{fig:teaser}
\end{figure}

\section{Introduction}
\label{sec:intro}

Fully automatic instrument segmentation supports endoscopic navigation and safety
monitoring~\cite{rueckert2024review,yao2026prototype,yue2024surgicalsam}, producing one mask for
every video frame without human interaction. Segment Anything Models~\cite{kirillov2023sam,carion2026sam3} provide a strong, reusable mask prior,
yet native SAM inference begins with user-supplied points or boxes. Direct deployment
therefore reduces to prompt acquisition: an endoscopic system must generate its own
point source from each frame.

Automatic endoscopic prompting must satisfy three coupled requirements: the points must
fall on the instrument foreground despite thin shafts, articulated tips, glare, occlusion,
and disconnected appearances; more than one configuration is valid, so several complete
hypotheses must be produced; and the choice among them must be ground-truth (GT) free.

For us the third requirement binds: the model must choose its own points.
Fine-tuning a large
segmentation model is data intensive~\cite{ma2024medsam,zhang2023samed}. A larger source set can
produce strong source-domain accuracy, yet the adapted response may still degrade
on a new dataset from the same clinical
domain~\cite{haralovic2026adaptation,chattopadhyay2026imprecise}. Freezing
the foundation pathway and learning only a compact point source lowers that demand and
leaves SAM's point-prompt interface untouched.
Attribution imposes a fourth requirement: the response function must stay fixed,
which freezing delivers. Under
a rigid $(+64,+64)$-pixel shift of the GT-derived reference (one eighth
of the canvas width), the Dice-point drop is 60.3 with a frozen decoder but 3.8 under LoRA~\cite{hu2022lora} fine-tuning: the response
stays sensitive to coordinate error, so a change in
decoded accuracy is attributable to the point
source~\cite{haralovic2026adaptation}. Adapting the decoder moves the
response function with the training signal; freezing it is what makes the
\emph{localization gap} measurable, the distance from the strongest zero-training rule
to a GT-derived point reference. Our frozen-pathway protocol measures this
gap.

Existing automatic prompting predicts a single point set or mask from image features~\cite{wu2023selfprompt,chen2025aopsam,zhou2024samsp,xie2025selfpromptsam,zhang2025hspsam},
transfers prompts from references or concepts~\cite{zhang2024persam,zhou2026m2c,jiang2026medsam3},
refines supplied geometry~\cite{nuren2025rpsam2,meyer2026s4m}, or adapts the prompt
encoder~\cite{shaharabany2023autosam,he2024apseg}. ProSAM samples probabilistic
prompt embeddings~\cite{wang2025prosam} and SeqSAM predicts mask hypotheses
autoregressively~\cite{towle2025seqsam}; both aggregate in latent or mask
space. No single line supplies all three requirements; fully automatic
endoscopy needs them in one point-set module that returns explicit coordinates.

We propose \textbf{SetPlanner}, a plug-in point-set planner with 1.52M parameters.
Frozen image features condition a flow that samples complete unordered
$K$-point sets. Fig.~\ref{fig:teaser} contrasts such complete sets with a
jittered score map, whose repeated draws are nearly identical. The native SAM pathway decodes eight candidates that share a single
image encoding, a
consensus readout returns one observed candidate, and disagreement supplies
a failure score that flags frames for review. On Kvasir-Instrument, SetPlanner recovers 96\%
of a 44.4-point \emph{localization gap} under a single frozen SAM. Across Kvasir, Endoscapes,
and EndoVis, it wins all six transfer routes over a LoRA-adapted system, which uses random point
prompts. We make three contributions:
\begin{list}{$\bullet$}{
  \setlength{\leftmargin}{1.6em}
  \setlength{\labelwidth}{0.6em}
  \setlength{\labelsep}{0.35em}
  \setlength{\itemindent}{0pt}
  \setlength{\itemsep}{0pt}
  \setlength{\parsep}{0pt}
  \setlength{\topsep}{0pt}
  \setlength{\partopsep}{0pt}}
\item We formulate automatic endoscopic prompting as a point-set planning problem
and define a frozen-pathway protocol that isolates the point source. A frozen
decoder keeps the response coordinate-sensitive, so the localization gap is measurable at low
data cost.
\item \textbf{SetPlanner} plans complete unordered $K$-point sets
with a permutation-aware conditional flow and geometry-aware targets,
so an ambiguous frame yields several distinct coherent hypotheses.
\item We reduce a decoded candidate pool to one mask by a GT-free consensus readout and
reuse the candidates' disagreement as a test-time failure score.
\end{list}

\begin{figure*}[t]
\centering
\includegraphics[width=\textwidth]{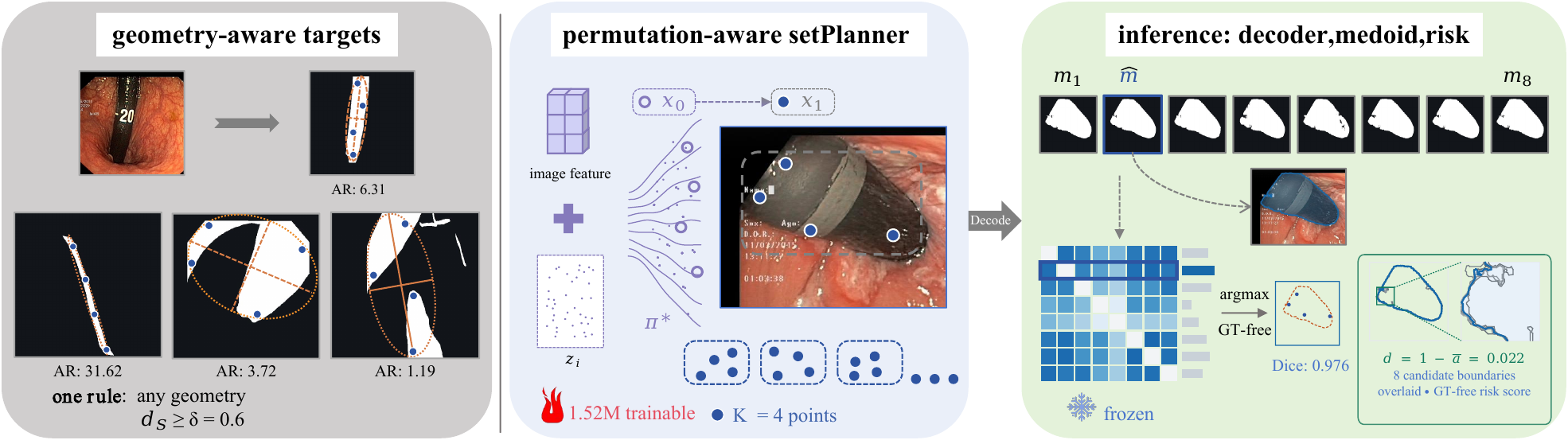}
\caption{\small \textbf{SetPlanner, left to right.} Targets $x_1$ come from the eroded
foreground under a Mahalanobis floor in the foreground's second-moment frame;
AR = axis ratio. In model coordinates $[-1,1]$, the optimal assignment
$\pi^\star$ pairs a Gaussian draw with the target set. $M{=}8$ decoded
candidates reduce to one by a consensus readout; disagreement $d$ ranks failures
without GT.}
\label{fig:pipeline}
\end{figure*}

\section{Method}
\label{sec:method}

Planning therefore means choosing a point set $P\in\mathbb{R}^{K\times2}$ from the
image alone, with no access to the GT mask $Y$ at test time.
SetPlanner adds one trainable component to an otherwise frozen SAM: a 1.52M-parameter
point-set planner that supplies the prompts the native pathway expects.
Fig.~\ref{fig:pipeline} follows one frame through it. A \emph{point set} holds $K$ points;
a \emph{candidate pool} holds $M$ decoded sets.

\subsection{Frozen-pathway localization gap}
\label{ssec:gap}
Under our frozen-pathway protocol, $f_\phi$ is the frozen SAM mapping
from image $I$ and point set $P\in\mathbb{R}^{K\times 2}$ to a mask, with GT
mask $Y$. We
prescribe $\mathcal{P}_0$ as five zero-training rules in the style of
training-free prompt strategies~\cite{yuan2026guidelines}: uniform,
center-Gaussian, color (lowest chroma span), desaturation
(lowest saturation), and specular. On Kvasir the color rule is
the strongest of the five, so Table~\ref{tab:points} groups it with the two rules
that use no image information. With $f_\phi$ and $K{=}4$ fixed, the
\emph{localization gap} for this protocol is
\begin{equation}
\label{eq:hole}
\begin{aligned}
\Delta_{\mathrm{loc}}
&=100\cdot\Big(\mathbb{E}\big[\Dice(f_\phi(I,P^\star),Y)\big]\\
&\quad-\max_{p\in\mathcal{P}_0}
\mathbb{E}\big[\Dice(f_\phi(I,p(I)),Y)\big]\Big),
\end{aligned}
\end{equation}
where $P^\star$ is sampled from the eroded GT foreground through the same
point-prompt interface.

The protocol fixes two reference levels: the rule floor (the best of $\mathcal{P}_0$) and
the GT-derived reference $P^{\star}$. Their difference is the \emph{localization gap}: the
accuracy available from improving the point source while the decoder and interface remain
fixed. On Kvasir it is $\Delta_{\mathrm{loc}}{=}44.4$ Dice points. Learned point sources
sit between the two levels (Table~\ref{tab:points}).
\begin{table}[t]
\caption{\small \textbf{Automatic point sources on frozen SAM.} Kvasir. Hit is the
fraction of points inside GT. Best scores are bold.}
\label{tab:points}
\centering
\renewcommand{\arraystretch}{1.00}
\small
\setlength{\tabcolsep}{3.2pt}
\begin{tabular*}{\columnwidth}{@{\extracolsep{\fill}}lccc@{}}
\toprule
Point source & Dice $\uparrow$ & Mask IoU $\uparrow$ & Hit $\uparrow$ \\
\midrule
\multicolumn{4}{@{}l}{\textit{Zero-training rules}} \\
\quad Uniform random & 0.091 & 0.060 & 0.114 \\
\quad Center Gaussian & 0.121 & 0.091 & 0.131 \\
\quad Color rule & 0.509 & 0.436 & 0.564 \\
\midrule
\multicolumn{4}{@{}l}{\textit{Learned sources ($M{=}1$, control), frozen decoder}} \\
\quad Logistic-1 & 0.922 & 0.873 & 0.970 \\
\quad Deterministic slots-1 & 0.924 & 0.880 & \textbf{0.975} \\
\quad SetPlanner-1 & 0.925 & 0.880 & 0.955 \\
\midrule
\multicolumn{4}{@{}l}{\textit{SetPlanner ($M{=}8$), frozen decoder}} \\
\quad \textbf{SetPlanner-8} & \textbf{0.934} & \textbf{0.894} & \textbf{0.975} \\
\bottomrule
\end{tabular*}
\end{table}

\subsection{SetPlanner: a better point-set planner}
\label{ssec:setplanner}
Endoscopic tool geometry couples the positions inside each point set: a useful
hypothesis must span a thin, oriented foreground despite glare
and partial occlusion. SetPlanner therefore generates each set jointly rather than
point by point, so that repeated draws differ by more than perturbing one prediction can.
Training targets $x_1$ are sampled from the foreground of $Y$ eroded by $4$\,px. Their pairwise
separation is a dimensionless Mahalanobis distance in the foreground's second-moment
frame, normalized by principal-axis lengths and constrained above
$\delta{=}0.6$. This geometry-aware constraint distributes points along elongated tools
while adapting to scale and orientation. Target sets are drawn by rejection sampling;
if no feasible set is found, $\delta$ is halved and then dropped. Gaussian noise $x_0$ supplies the residual spatial ambiguity.
A rectified flow~\cite{liu2023rectified,lipman2023flow} transports $x_0$ to $x_1$ with a
velocity field $v_\theta$ that SetPlanner implements as four 128-d blocks. Frozen SAM
features condition the flow: a $1{\times}1$ convolution maps the image encoder's
$256\times64\times64$ output to 128 channels and average-pools it to $16{\times}16$
key--value tokens, which the $K$ point tokens attend to in each block by cross-attention.
A sinusoidal embedding of $t$, added to a global token pooled from them, drives adaptive
layer norm in every block. Point-token self-attention captures within-set interactions, and
coordinate-only point tokens with shared parameters
make $v_\theta$ permutation equivariant~\cite{li2025unorderedflow}.
Each integration yields one coherent unordered set.

\subsection{Permutation-aware training and inference}
\label{ssec:objective}
The target carries no point identities, so the objective must treat every
permutation of a set identically. With
$x_0\sim\mathcal{N}(0,\mathbf{I}_{K\times2})$ and $x_1$ (eroded-foreground
targets) as defined in Sec.~\ref{ssec:setplanner}, and $t\sim\mathcal{U}(0,1)$,
we solve the noise-to-target correspondence $\pi\in S_K$, one of the $K!$ assignments,
before regression:
\begin{equation}
\label{eq:rfloss}
\begin{aligned}
\pi^\star&=\arg\min_{\pi\in S_K}\sum_{k=1}^{K}
\|x_{1,\pi(k)}-x_{0,k}\|_2^2,\\
 x_t&=(1-t)x_0+t\pi^\star(x_1),\\
\mathcal{L}(\theta)&=\mathbb{E}\left\|
 v_\theta(x_t,t,c)-\bigl(\pi^\star(x_1)-x_0\bigr)\right\|_2^2,
\end{aligned}
\end{equation}
where $c$ is the SAM image feature. SetPlanner's $\theta$ holds all trainable
parameters; the SAM pathway stays frozen. At inference, $M{=}8$ independent noise
draws are each integrated from $t=0$ to $t=1$ by Euler in eight steps, producing $M$
complete point sets.

\subsection{Candidate selection and failure ranking}
\label{ssec:select}
SAM decodes $M{=}8$ candidates; let $\{m_i\}_{i=1}^{M}$ be their masks --- a common
output space in which to compare complete point-set hypotheses. Deployment requires
one mask per frame, and the choice among them must be GT-free, resting only on model
responses available at test time. Our
\emph{consensus readout} takes the medoid, the candidate agreeing most with the
others. It returns $m_{\hat{\imath}}$, one observed candidate, and so keeps the
selected mask attributable to a point set. We use candidate disagreement $d$ as a
failure score~\cite{chowdhury2026bald,gupta2026crossmodel}:
\begin{align}
\label{eq:medoid}
\hat{\imath}
&=\arg\max_i\tfrac{1}{M{-}1}\sum\nolimits_{j\neq i} a(m_i,m_j),\\
\label{eq:disagree}
d&=1-\tfrac{2}{M(M{-}1)}\sum\nolimits_{i<j} a(m_i,m_j),
\end{align}
where $a$ is the pair IoU, the intersection-over-union overlap between two masks.
Under the frozen decoder, independent noise draws make $d$ a measure of
variation induced by the automatic point source. The readout is defined for any
$M\geq2$ and is applied unchanged to every candidate pool we compare.

\section{Experiments}
\label{sec:exp}

\subsection{Protocol}
\label{ssec:protocol}
The experiment uses the official Kvasir-Instrument 472/118
split~\cite{jha2021kvasir}, a $512{\times}512$ canvas, and native-resolution Dice. Every
Dice value is a union-mask Dice and a three-seed mean: instruments in a frame are merged.
Pair IoU averages the overlap among candidates in
one pool. Table~\ref{tab:points} also reports Mask IoU on the same masks and Hit,
the fraction of points inside GT. The SAM image encoder, prompt
encoder, and decoder remain frozen.

Candidate pools
use $M{=}8$ candidates decoded from point sets of $K{=}4$ points; selection saturates beyond this
budget, and the plug-in adds $1.0\%$ FLOPs over a single prompt set. Stochastic sources average repeated sampling.
SetPlanner is compared with two automatic point sources, both instantiated at the same
point-prompt interface:
a 257-parameter logistic head on the frozen embedding, in the style of
Self-Prompting~\cite{wu2023selfprompt}, whose eight sampled sets are reduced
by the same readout (Logistic); and a SAM adapted with LoRA on the same
source data, prompted with random points inside the endoscope field of view.
Both return explicit coordinates, so each source's error is observable.
A deterministic-slots control provides the $M{=}1$ baseline: the same planner with $K$
learnable slots in place of the noise input, regressing one $K$-point set without flow sampling.
Local perturbations use $\sigma{=}0.02$ in model coordinates.
Each candidate is decoded once, resampled to $512{\times}512$, and thresholded at 0.5.

Each dataset row retrains the same planner: Endoscapes uses a video-grouped 334/112
split~\cite{mascagni2025endoscapes}; EndoVis, 1778 nonempty training frames and the
official 1200-frame test set~\cite{allan2019endovis}, half of it
temporally adjacent to training video. GT-derived references use an isotropic $48$\,px minimum
separation on Endoscapes and an anisotropic $\delta{=}0.6$ on Kvasir and EndoVis.
We first test that the frozen response stays coordinate-sensitive: a rigid $(+64,+64)$ shift
of the GT-derived reference costs 60.3 Dice points under the frozen decoder, 3.8 under
LoRA, and 0.6 under a fully fine-tuned decoder, so decoded accuracy stays attributable to
the point source.

\subsection{Gap recovery, reuse, and transfer}
\label{ssec:three}

Table~\ref{tab:all} compares the three systems on every training set. Its top block fixes the
protocol's two reference levels and the gap between them; the lower block reports Dice, and its
last column $\Delta_{\mathrm{m}}$ is the pool's gain over one candidate. SetPlanner reaches
0.934 against a 0.953 GT-derived reference,
recovering 96\% of the localization gap; the remaining 1.9 points are prompt-source
headroom.

The same planner is then retrained per dataset and reused
across backbones. In-domain, it improves on Logistic by 3.0 points on Kvasir
with SAM1 and 1.2 on EndoVis, and on Kvasir and Endoscapes the LoRA-adapted system
reaches only 0.703 and 0.693 Dice, whereas the frozen
plug-in stays above 0.86 on all three: model adaptation needs a large source set, the
frozen plug-in does not. After training on EndoVis, the LoRA-adapted system transfers to only
0.751 on Kvasir
and 0.803 on Endoscapes, where frozen SetPlanner reaches 0.818 and 0.848. The frozen plug-in
system wins all six transfer routes and raises mean Dice from 0.695 to 0.839. The pool's gain
over one candidate also depends on the decoder: it survives a frozen decoder but collapses once
LoRA adapts it, since an adapted decoder absorbs the variation between point sets.

\begin{table}[t]
\caption{\small \textbf{SetPlanner against Logistic and LoRA,
all sharing a frozen image encoder.} All three systems decode an $M{=}8$ candidate pool and reduce it
with the same consensus readout. Top: per-target reference levels. Bottom: Dice by training set;
each block tests on one target, so its first row is in-domain and the rows below it
transfer. Bold marks the best system in a row.
$\Delta_{\mathrm{m}}$ is the eight-candidate gain
over one candidate, in the frozen/LoRA order.}
\label{tab:all}
\centering
\small
\renewcommand{\arraystretch}{0.94}
\setlength{\tabcolsep}{1.2pt}

\begin{tabular*}{\columnwidth}{@{\extracolsep{\fill}}lcccc@{}}
\toprule
& Kvasir & Endoscapes & EndoVis & Kvasir (SAM1) \\
\midrule
rule floor & 0.509 & 0.181 & 0.198 & 0.594 \\
GT-ref & 0.953 & 0.901 & 0.929 & 0.937 \\
\rowcolor{blue!10}%
$\Delta_{\mathrm{loc}}$ & 44.4 & 72.1 & 73.0 & 34.3 \\
\bottomrule
\end{tabular*}
\vspace{-0.6\baselineskip}
\begin{tabular*}{\columnwidth}{@{\extracolsep{\fill}}lr@{\quad}rrr@{}}
& \multicolumn{1}{c}{SetPlanner} & \multicolumn{1}{c}{Logistic} & \multicolumn{1}{c}{LoRA} & \multicolumn{1}{c}{$\Delta_{\mathrm{m}}$} \\
\midrule
\textit{Kvasir} & & & & \\
\quad Kvasir 472 & \textbf{0.934} & 0.933 & 0.703 & $+0.9$/$+0.0$ \\
\quad Endoscapes 334 & \textbf{0.888} & \NA & 0.520 & $+4.3$/$+0.8$ \\
\quad EndoVis 1778 & \textbf{0.818} & \NA & 0.751 & $+5.1$/$+1.0$ \\
\midrule
\textit{Kvasir (SAM1)} & & & & \\
\quad Kvasir 472 & \textbf{0.900} & 0.870 & \NA & \NA \\
\midrule
\textit{Endoscapes} & & & & \\
\quad Endoscapes 334 & 0.862 & \textbf{0.870} & 0.693 & $+4.1$/$+0.1$ \\
\quad Kvasir 472 & \textbf{0.803} & \NA & 0.582 & $+4.6$/$+1.1$ \\
\quad EndoVis 1778 & \textbf{0.876} & \NA & 0.752 & $+7.7$/$+1.9$ \\
\midrule
\textit{EndoVis} & & & & \\
\quad EndoVis 1778 & 0.911 & 0.899 & \textbf{0.929} & $+3.5$/$+0.1$ \\
\quad Kvasir 472 & \textbf{0.801} & \NA & 0.761 & $+8.2$/$+1.3$ \\
\quad Endoscapes 334 & \textbf{0.848} & \NA & 0.803 & $+4.2$/$+0.5$ \\
\bottomrule
\end{tabular*}
\end{table}

\subsection{Ablations and failure ranking}
\label{ssec:law}
Generation and selection both prove load-bearing: one decides what
the pool contains, the other what is read from it.
Local perturbations produce near-duplicate pools with pair IoU at least
0.965, whereas SetPlanner lowers it to 0.951 and reaches 0.934 selected Dice
(Table~\ref{tab:spacing}): generation, not perturbation, is what makes a pool
worth selecting from. Eight
independently trained models reach 0.929, 0.5 points below SetPlanner at eight times
the parameter count, so the
advantage does not come from capacity. Their pools are near-duplicates at
pair IoU 0.964, so diversity does not come from ensembling either. The
points-per-set rows peak at $K{=}4$, and a larger per-set budget leaves less for
the pool to add.

\begin{table}[t]
\caption{\small \textbf{Candidate generation and selection ablations.} Kvasir.
$M{=}1$ is one candidate, $M{=}8$ the medoid-selected pool, and Oracle the best;
all report Dice. The candidate-source rows perturb deterministic
slots; the last group varies the selector on one pool.
TTA = test-time augmentation. Predicted IoU selects by SAM's own mask-quality score.}
\label{tab:spacing}
\centering
\renewcommand{\arraystretch}{0.88}
\small
\setlength{\tabcolsep}{2.3pt}
\begin{tabular*}{\columnwidth}{@{\extracolsep{\fill}}lcccc@{}}
\toprule
Variant & $M{=}1$ & $M{=}8$ & Oracle & Pair IoU \\
\midrule
\multicolumn{5}{@{}l}{\textit{Points per set}} \\
\quad $K{=}1$ & 0.855 & 0.911 & 0.932 & 0.865 \\
\quad $K{=}2$ & 0.912 & 0.928 & 0.941 & 0.936 \\
\quad $K{=}4$ & 0.925 & \textbf{0.934} & \textbf{0.947} & 0.951 \\
\quad $K{=}8$ & 0.909 & 0.918 & 0.932 & 0.949 \\
\midrule
\multicolumn{5}{@{}l}{\textit{Candidate source}} \\
\quad 8 independent models & \NA & 0.929 & 0.942 & 0.964 \\
\quad Deterministic slots & 0.924 & \NA & \NA & \NA \\
\quad $+$ flip (TTA) & 0.924 & 0.923 & 0.936 & 0.970 \\
\quad $+$ cond. noise & 0.923 & 0.923 & 0.925 & 0.997 \\
\quad $+$ slot noise & 0.923 & 0.923 & 0.925 & 0.996 \\
\quad $+$ point jitter & 0.924 & 0.926 & 0.934 & 0.965 \\
\midrule
\multicolumn{5}{@{}l}{\textit{Selector on the SetPlanner pool}} \\
\quad Random & \NA & 0.929 & 0.947 & 0.951 \\
\quad Predicted IoU & \NA & 0.924 & 0.947 & 0.951 \\
\quad \textbf{Medoid} & \NA & \textbf{0.934} & 0.947 & 0.951 \\
\bottomrule
\end{tabular*}
\end{table}

Medoid selection is the best of the three selectors. Candidate disagreement ranks Kvasir frames with
Dice$<0.7$ at an area under the ROC curve (AUROC) of 0.969 and provides a test-time failure
score~\cite{borges2026softdice}, so frames can be deferred by rank.

\section{Conclusion}
\label{sec:conc}
SetPlanner supplies the automatic point source that frozen SAM needs for endoscopic
video. Automatic prompting is formulated as point-set planning under a
frozen-pathway protocol. This plug-in preserves the point-prompt interface, plans
complete unordered hypotheses from geometry-aware targets, and selects among them with a GT-free consensus readout whose
disagreement also ranks failures. Its disagreement score reaches AUROC 0.969 on low-Dice
cases, and SetPlanner wins all six transfer routes. One frozen architecture therefore delivers low data demand
and portability across backbones and datasets; under the frozen-pathway
protocol it recovers 96\% of the localization gap without touching the decoder. Absolute Dice and the gap size remain dataset-
and backbone-specific.

\clearpage
{\small
\bibliographystyle{IEEEbib}
\bibliography{paper}}

\section*{Compliance with Ethical Standards}
This study used only openly available, de-identified human subject data from
Kvasir-Instrument~\cite{jha2021kvasir}, Endoscapes~\cite{mascagni2025endoscapes},
and EndoVis2017~\cite{allan2019endovis}. Ethical approval was not required, as
confirmed by the licenses under which these open-access datasets are distributed.

\end{document}